\documentclass[runningheads]{llncs}
\usepackage{graphicx}
\usepackage{ragged2e} 
\usepackage{booktabs,makecell, multirow, tabularx}
\usepackage{fontawesome}
\begin{document}
%
\title{Structure Diagram Recognition in Financial Announcements}
\titlerunning{Structure Diagram Recognition in Financial Announcements}
%
\author{Meixuan Qiao\inst{1} \and
Jun Wang\inst{2}(\faEnvelopeO) \and
Junfu Xiang\inst{2} \and
Qiyu Hou\inst{2} \and
Ruixuan Li\inst{1}
}
\authorrunning{M. Qiao et al.}
%
\institute{Huazhong University of Science and Technology \\
\email{\{qiaomeixuan,rxli\}@hust.edu.cn}
\and
iWudao Tech\\ 
\email{\{jwang,xiangjf,houqy\}@iwudao.tech}}
\maketitle              

\begin{abstract}

Accurately extracting structured data from structure diagrams in financial announcements is of great practical importance for building financial knowledge graphs and further improving the efficiency of various financial applications. 
First, we proposed a new method for recognizing structure diagrams in financial announcements, which can better detect and extract different types of connecting lines, including straight lines, curves, and polylines of different orientations and angles. 
Second, we developed a semi-automated, two-stage method to efficiently generate the industry's first benchmark of structure diagrams from Chinese financial announcements, where a large number of diagrams were synthesized and annotated using an automated tool to train a preliminary recognition model with fairly good performance, and then a high-quality benchmark can be obtained by automatically annotating the real-world structure diagrams using the preliminary model and then making few manual corrections.
Finally, we experimentally verified the significant performance advantage of our structure diagram recognition method over previous methods.

\keywords{Structure Diagram Recognition  \and Document AI \and Financial Announcements}
\end{abstract}
\section{Introduction}

As typical rich-format business documents, financial announcements contain not only textual content, but also tables and graphics of various types and formats, which also contain a lot of valuable financial information and data. 
Document AI, which aims to automatically read, understand, and analyze rich-format business documents, has recently become an important area of research at the intersection of computer vision and natural language processing~\cite{DocumentAI2021}. 
For example, layout analysis~\cite{DocumentAI2021}, which detects and identifies basic units in documents (such as headings, paragraphs, tables, and graphics), and table structure recognition~\cite{DocumentAI2021}, which extracts the semantic structure of tables, have been widely used in the extraction of structured data from various rich-format documents, including financial announcements, and a number of benchmark datasets have been established~\cite{DocBank2020,PubLayNet2019,PubTabNet2019,GTE2020}. 
Nevertheless, research on the recognition and understanding of graphics in rich-format documents remains in the early stages. Although some studies have focused on recognizing flowcharts~\cite{frankdas-DAS2018,Arrow-RCNN2021,Sketch2BPMN2021,DiagramNet2021}, statistical charts~\cite{Chart-to-Text-ACL2022,DeMatch2021,chart-ICDAR2021,LineChart-mdpi-2021,LineChart-WACV-2022} and geometry problem~\cite{seo2014diagram,chen2021geoqa,Lu2021InterGPSIG}, to the best of our knowledge, there is no existing work explicitly dedicated to recognizing and understanding various types of structure diagrams in financial announcements.

\begin{figure}
\centering
\includegraphics[width=\textwidth]{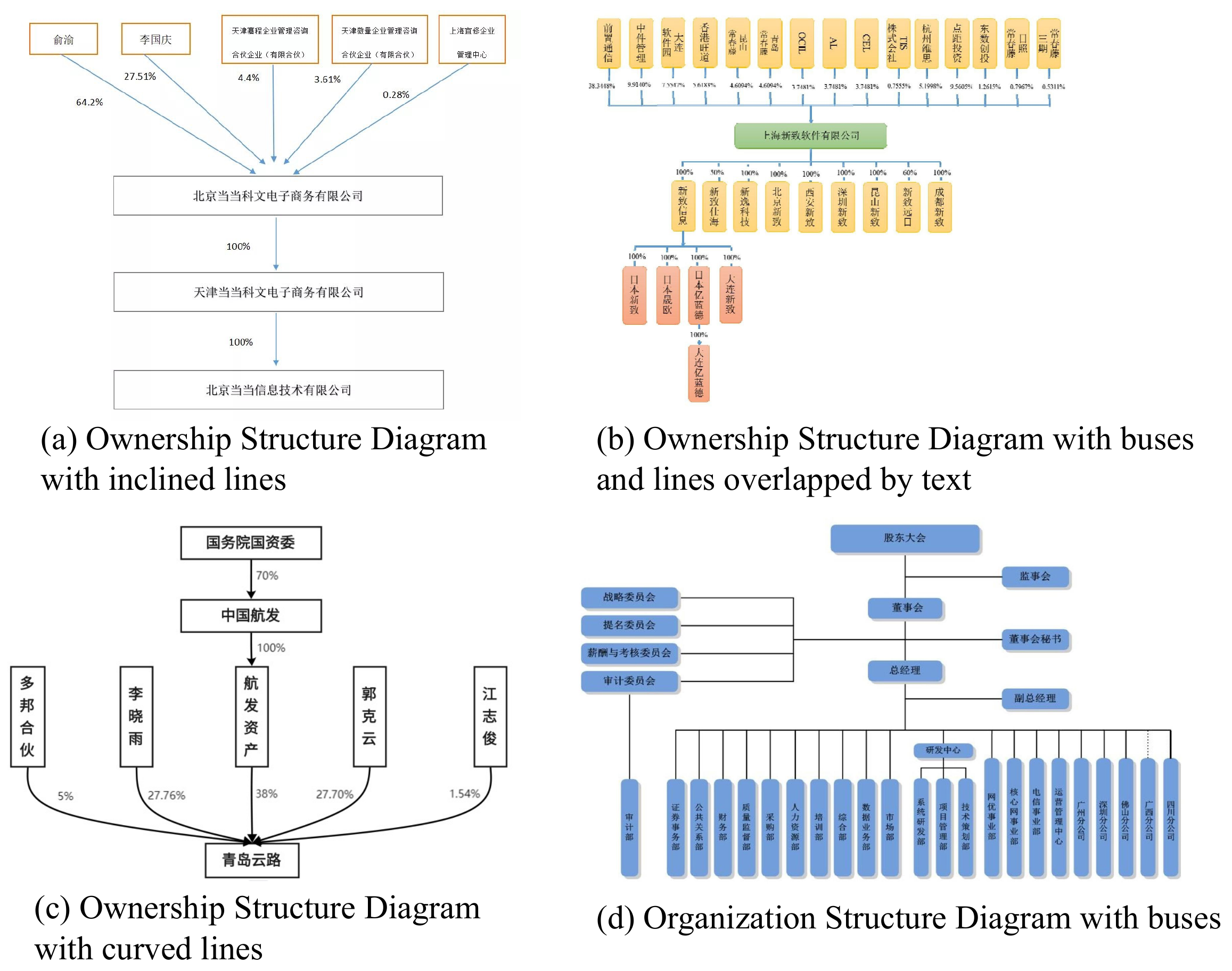}
\caption{Typical examples of Structure Diagrams in Chinese Financial Announcements} \label{figure-ownership-organization-examples}
\end{figure}

Figure~\ref{figure-ownership-organization-examples} (a), (b), and (c) are typical ownership structure diagrams extracted from Chinese financial announcements, where the nodes represent institutional or individual entities, and the connecting lines represent the ownership relationship and proportion between a pair of entities.
Although the ownership structure information can also be obtained from a business registration database, generally there is a timeliness gap, and it often takes some time for the latest ownership changes disclosed in financial announcements to be updated and reflected in the business registration database. 
Figure~\ref{figure-ownership-organization-examples} (d) shows a typical organization structure diagram in a Chinese financial announcement, where the nodes represent the departments or occupations in an institution, and the connecting lines between the nodes represent the superior-subordinate relationships. 
Recognizing organization structure diagrams can help understand the decision-making structure of an institution and analyze its business planning for future development. 
%
Currently, many practitioners in the financial industry still rely on manual data extraction from structure diagrams in Chinese financial announcements, which is time-consuming and error-prone. 
Therefore, it is of great practical significance to improve the efficiency of various related financial applications by automatically extracting structured data from structure diagrams in financial announcements and constructing corresponding financial knowledge graphs in a timely and accurate manner.

Although the recent deep learning-based object detection methods~\cite{frankdas-DAS2018,Arrow-RCNN2021,Sketch2BPMN2021,DiagramNet2021,FR-DETR-2022} have improved diagram recognition compared to traditional image processing-based methods~\cite{Rusinol-IR2014,morzinger-2012-visual,thean-2012-textual}, they mainly focus on flowchart recognition and are not very effective at structure diagram recognition, especially not good at detecting various connecting lines in structure diagrams.

First, to address the above problem, we proposed a new method called SDR by extending the Oriented R-CNN~\cite{Oriented-RCNN2021} with key point detection~\cite{Mask-RCNN2017}, which is particularly good at detecting various connecting lines with different orientations and angles, including straight lines, curves, and polylines.

Second, to overcome the lack of training data and the high cost of annotation, a two-stage method was developed to efficiently build a benchmark with high-quality annotations.
(1) An automated tool has been built to efficiently synthesize and annotate a large number of structure diagrams of different styles and formats, which can be used to train a preliminary structure diagram recognition model. 
(2) The preliminary model can automatically annotate real-world structure diagrams with very reasonable quality, so that only a very limited number of manual corrections are required.

Third, to evaluate the effectiveness of our methods in real scenarios, we used the above two-stage method to build the industry's first benchmark containing 2216 real ownership structure diagrams and 1750 real organization structure diagrams extracted from Chinese financial announcements, and experimentally verified the significant performance advantage of our SDR method over previous methods. 






\section{Related Works}\label{sec_related_work}

The work on flowchart recognition is the closest to the structure diagram recognition studied in this paper.
Early work mainly used traditional image processing methods based on Connected Components Analysis and various heuristic rules~\cite{Rusinol-IR2014,morzinger-2012-visual,thean-2012-textual}, which usually had poor performance in detecting dashed and discontinuous lines and were easily disturbed by noise on poor quality scanned images. 

Recently, some deep learning-based methods have been applied to flowchart recognition, and have made some progress compared to traditional image processing methods~\cite{frankdas-DAS2018,Arrow-RCNN2021,Sketch2BPMN2021,DiagramNet2021,FR-DETR-2022}.
Julca-Aguilar et al.~\cite{frankdas-DAS2018} first used Faster R-CNN~\cite{Faster-R-CNN2015} to detect symbols and connecting lines in handwritten flowcharts. 
Subsequently, Sch{\"a}fer et al. proposed a series of models~\cite{Arrow-RCNN2021,Sketch2BPMN2021,DiagramNet2021}, such as Arrow R-CNN~\cite{Arrow-RCNN2021}, to extend the Faster R-CNN based on the characteristics of handwritten flowcharts. 
The Arrow R-CNN, like the Faster R-CNN, is constrained to generating horizontal rectangular bounding boxes. Furthermore, Arrow R-CNN was initially developed to detect simple and direct connecting lines between nodes, and it treats the entire path between each pair of linked nodes as a single connecting line object. These two factors make it difficult to correctly detect certain connecting lines in the structure diagrams.
As shown in Figure~\ref{figure-cmp-inclined} (a), the horizontal bounding boxes of the two middle inclined lines are almost completely covered by the horizontal bounding boxes of the two outer inclined lines, resulting in one of the middle inclined lines not being correctly detected.
When nodes on different layers are connected via buses, as shown in Figure~\ref{figure-cmp-bus} (a), Arrow R-CNN can only annotate the entire multi-segment polyline connecting each pair of nodes as a single object, and causes the bounding boxes of the middle polylines to completely overlap with the bounding boxes of the outer polylines. And we can see that most of the connecting polylines containing the vertical short line segments in Figure~\ref{figure-cmp-bus} (a) are not detected.
Sun et al.~\cite{FR-DETR-2022} created a new dataset containing more than 1000 machine-generated flowcharts, and proposed an end-to-end multi-task model called FR-DETR by merging DETR~\cite{DETR-2020} for symbol detection and LETR~\cite{LETR-2021} for line segment detection. 
FR-DETR assumed that the connection path between each pair of connected symbols in flowcharts was all composed of straight line segments, and defined only straight line segments as the object of connecting line detection.
FR-DETR detects straight line segments by directly regressing the coordinates of the start and end points for each line segment, and thus can handle inclined line segments naturally. 
Although FR-DETR may be able to fit certain curves with straight lines, it cannot correctly detect connecting lines with high curvature. As shown in Figure~\ref{figure-cmp-curve} (a), all four curves are not detected by FR-DETR.
Also, in some ownership structure diagrams, connecting lines between nodes are overlapped with corresponding text, and this often causes FR-DETR to fail to detect these connecting lines with text. As shown in Figure~\ref{figure-cmp-overlap-text} (a), almost all of the short vertical lines overlaid with numbers are not detected.
FR-DETR's Transformer-based model also results in a much slower recognition speed. 
Our SDR method can not only better detect various connecting lines in structure diagrams, but also further parse the corresponding semantic structure and extract the structured data based on the obtained connecting relationships.

%

\section{System Framework}\label{sec_framework}


\subsection{Structure Diagram Detection} \label{diagram_detection}

\begin{figure}
\centering
\includegraphics[width=0.8\textwidth]{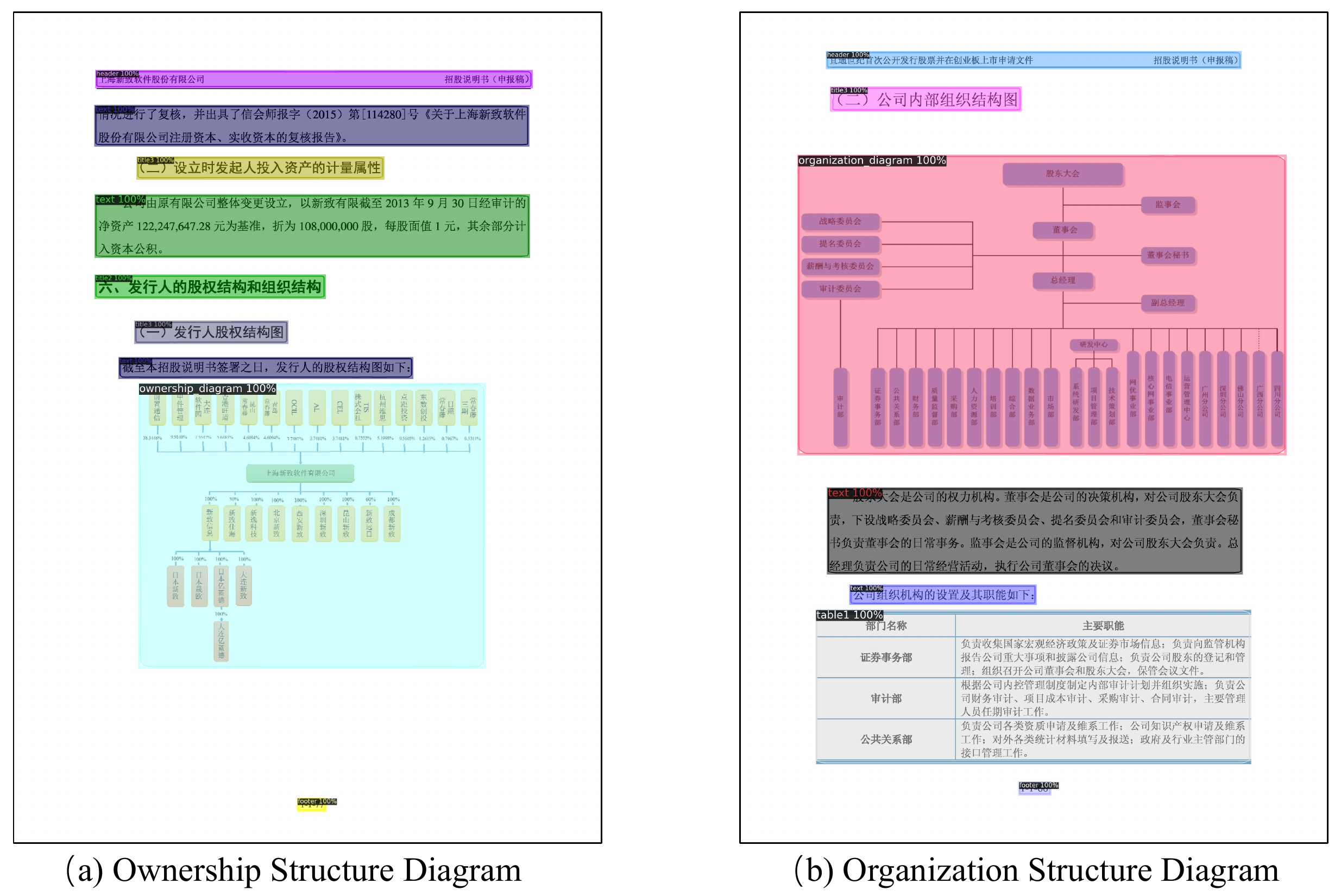}
\caption{Examples of structure diagram detection using layout analysis.} \label{figure-layoutanalysis}
\end{figure}

An ownership or organization structure diagram usually appears in a specific section of a financial announcement, so the text in the announcement, including the section titles, can be analyzed to determine the page range of the section where the structure diagram is located, and then the layout of the candidate page can be further analyzed to locate the bounding boxes of the required structure diagram. We used VSR~\cite{VSR2021} to implement layout analysis because it adds textual semantic features to better distinguish different types of diagrams with similar visual appearance (e.g., ownership and organizational structure diagrams) compared to models that use only visual features, such as Mask R-CNN~\cite{Mask-RCNN2017}, and has a much lower training cost compared to the pre-trained language models, such as LayoutLM~\cite{LayoutLMv2-2021,LayoutXLM2021,StructuralLM2021}.
The cyan area in Figure~\ref{figure-layoutanalysis} (a) and the pink area in Figure~\ref{figure-layoutanalysis} (b) are two examples of the ownership structure diagram and organization structure diagram, respectively, detected using layout analysis.

\subsection{Detection of connecting lines and nodes} \label{sec_detection_line_node}

\begin{figure}
\centering
\includegraphics[width=\textwidth]{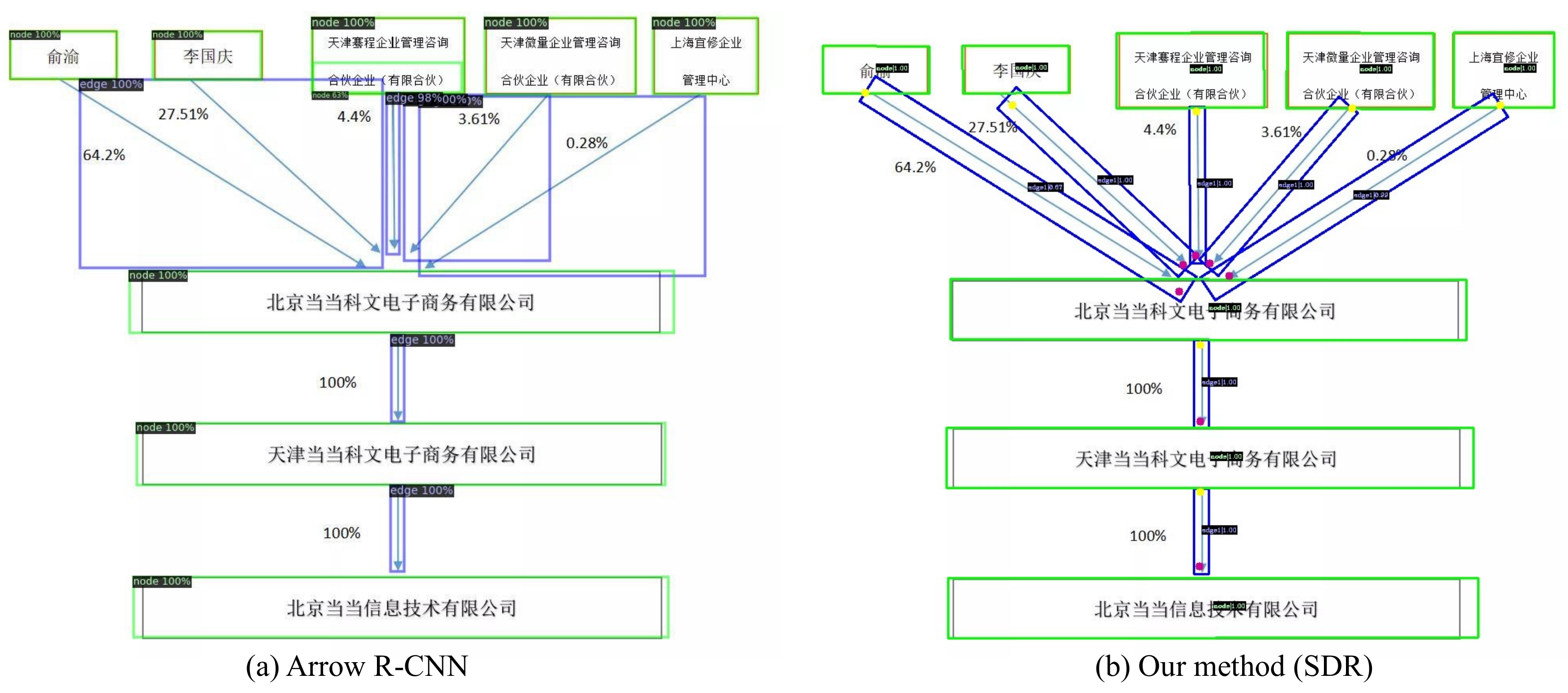}
\caption{Recognition of Ownership Diagram with inclined lines} \label{figure-cmp-inclined}
\end{figure}

\begin{figure}
\centering
\includegraphics[width=\textwidth]{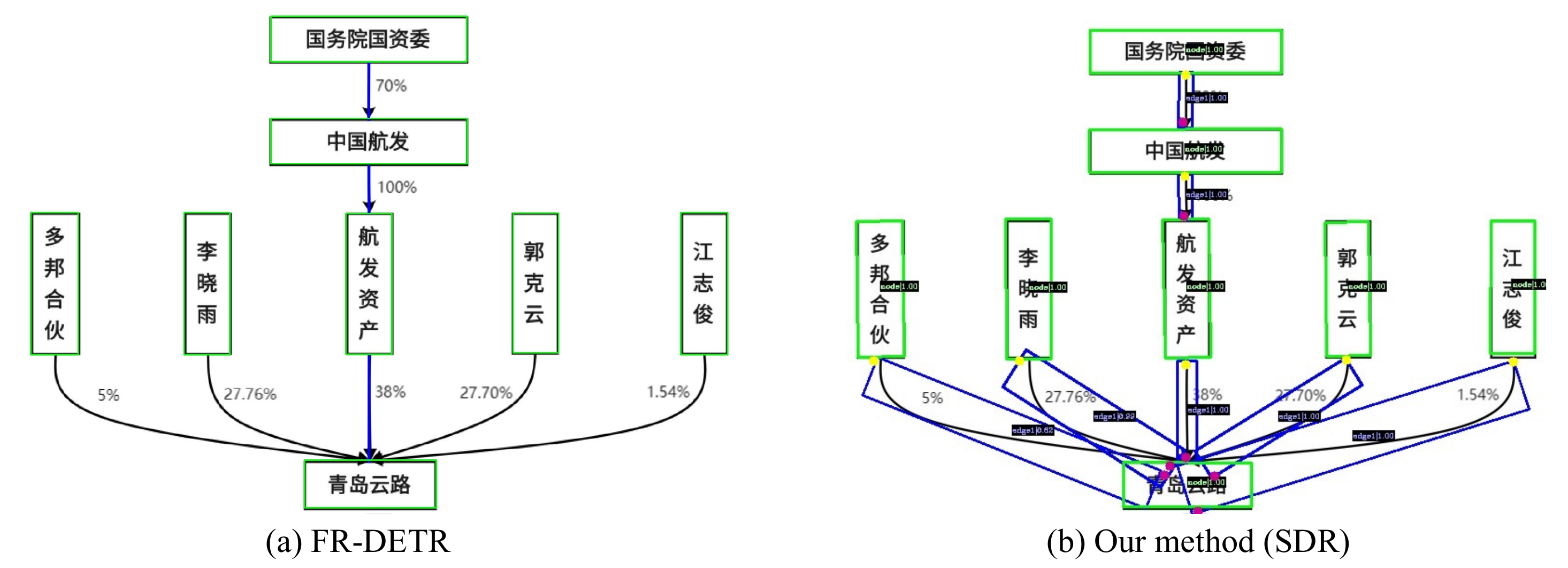}
\caption{Recognition of Ownership Diagram with curved lines} \label{figure-cmp-curve}
\end{figure}

\begin{figure}
\centering
\includegraphics[width=\textwidth]{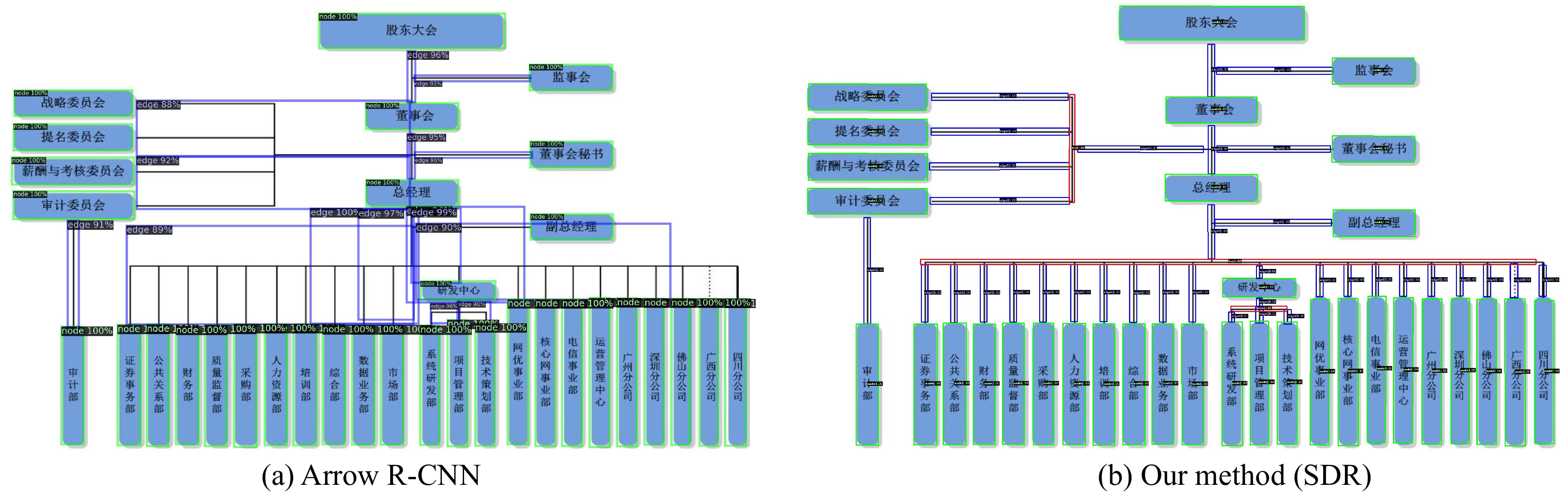}
\caption{Recognition of Ownership Diagram with bus structure} \label{figure-cmp-bus}
\end{figure}

\begin{figure}
\centering
\includegraphics[width=\textwidth]{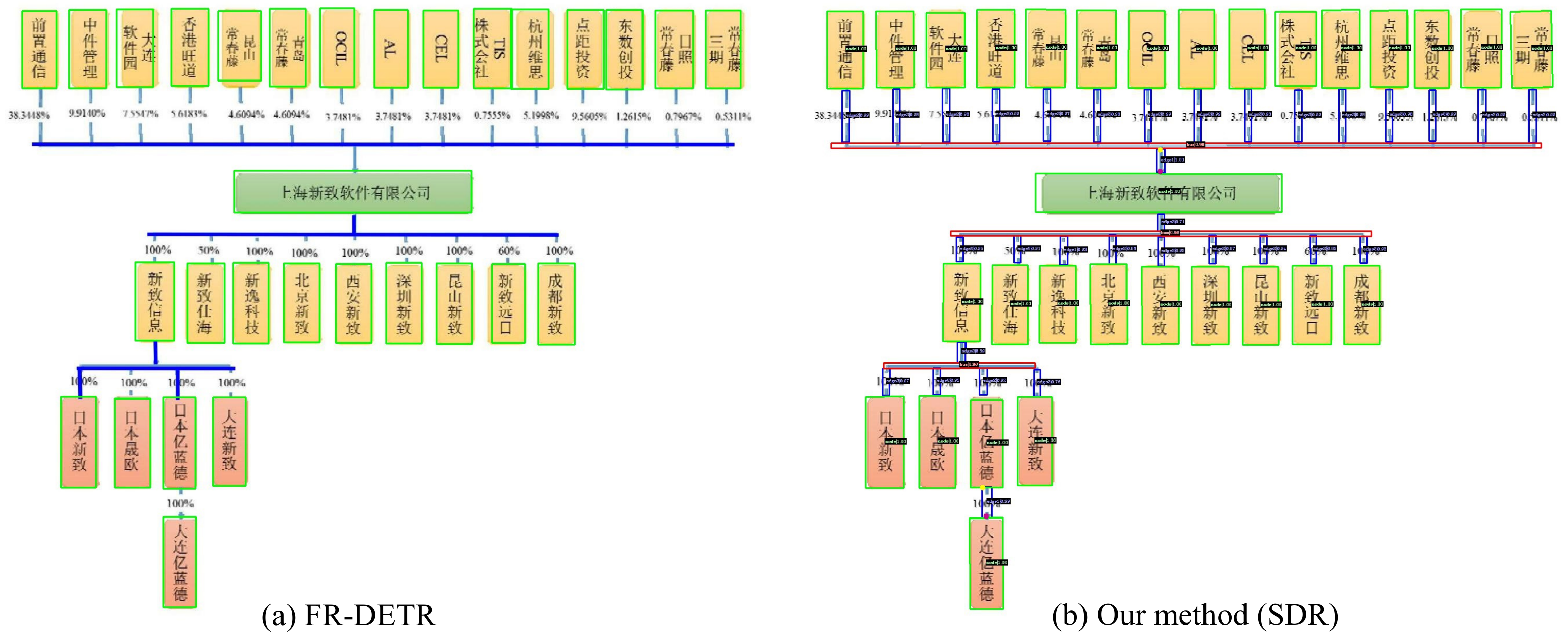}
\caption{Recognition of Ownership Diagram with lines overlapped by text} \label{figure-cmp-overlap-text}
\end{figure}

\begin{figure}
\centering
\includegraphics[width=\textwidth]{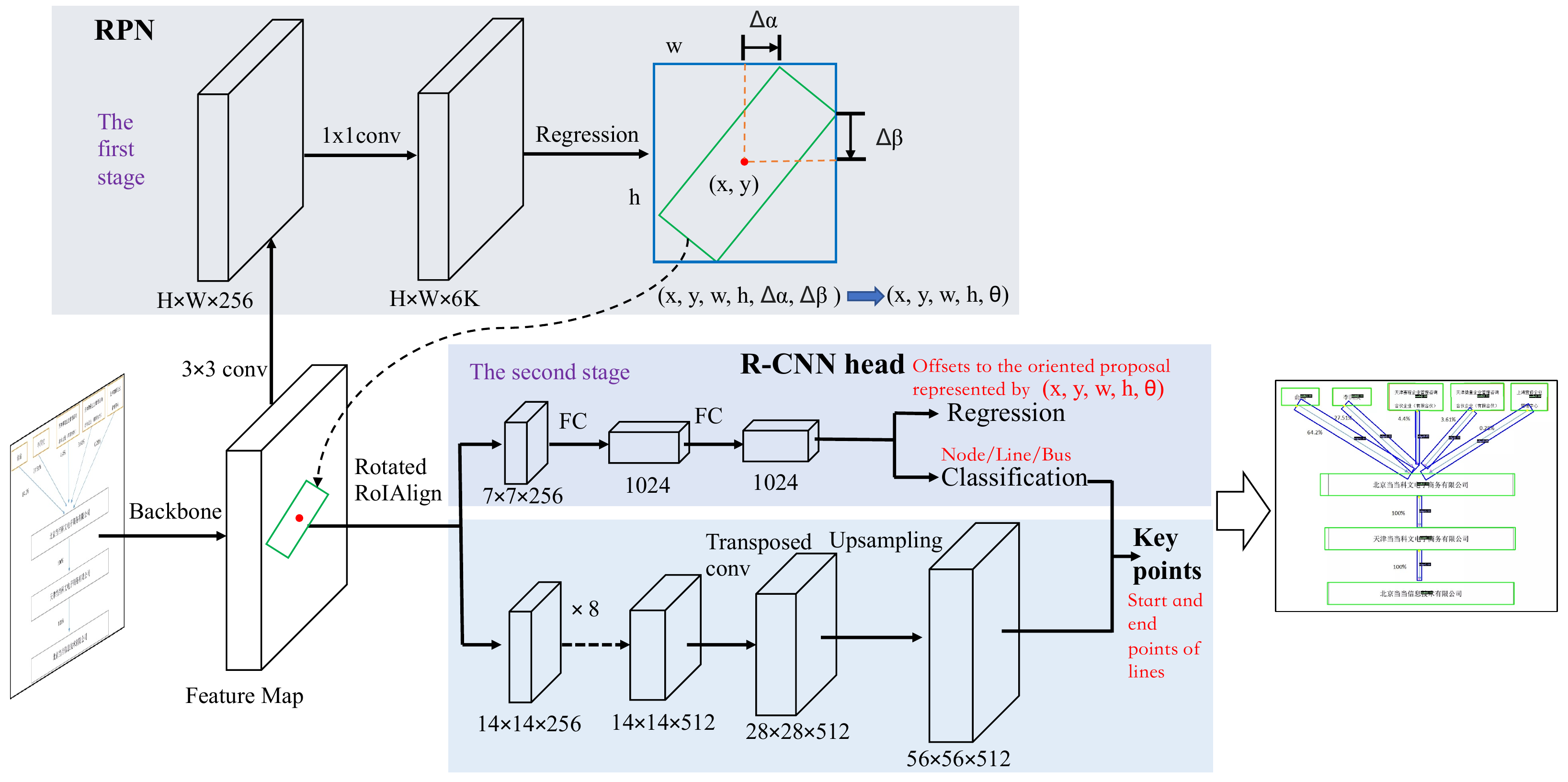}
\caption{Structure Diagram Recognition (SDR) Model} \label{Fig-network}
\end{figure}

\begin{figure}
\centering
\includegraphics[width=\textwidth]{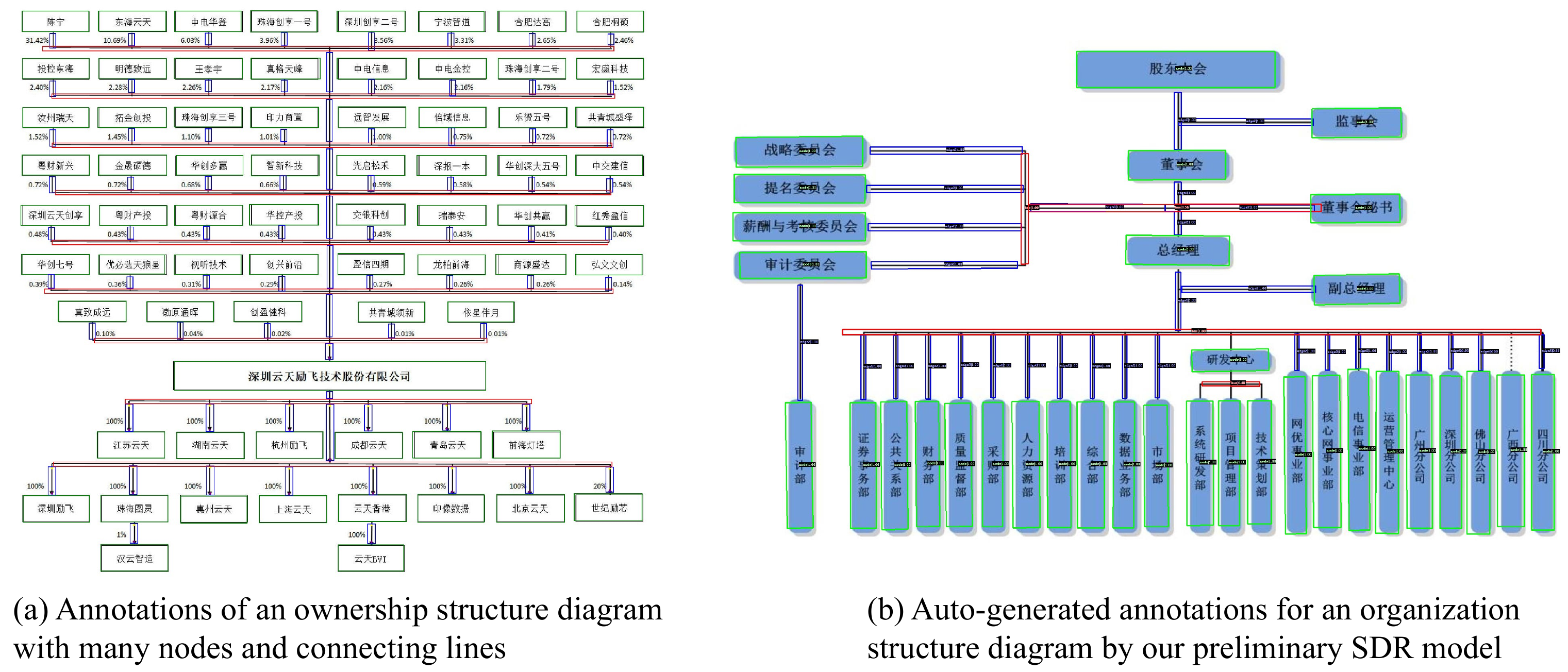}
\caption{Some annotation examples for structure diagrams} \label{figure-annotation-examples-1}
\end{figure}

This paper proposed a new method called SDR for structure diagram recognition, which can better handle various connection structures between nodes in particular. Figure~\ref{Fig-network} is an illustration of the network structure of our SDR model.
First, our SDR model extended the Oriented R-CNN model~\cite{Oriented-RCNN2021} to support the detection of oriented objects, and this allows the output bounding boxes to be rotated by a certain angle to achieve better detection of inclined or curved lines, as shown in Figure~\ref{figure-cmp-inclined} (b) and Figure~\ref{figure-cmp-curve} (b).
In the first stage of SDR, we extend the regular RPN in Faster R-CNN to produce oriented proposals, each corresponding to a regression output $(x,y,w,h,\Delta \alpha,\Delta \beta)$~\cite{Oriented-RCNN2021}. $(x, y)$ is the center coordinate of the predicted proposal, and $w$ and $h$ are the width and height of the external rectangle box of the predicted oriented proposal. $\Delta \alpha$ and $\Delta \beta$ are the offsets relative to the midpoints of the top and right sides of the external rectangle. Each oriented proposal generated by the oriented RPN is usually a parallelogram, which can be transformed by a simple operation into an oriented rectangular proposal represented by $(x,y,w,h,\theta)$.
The oriented rectangular proposal is projected onto the feature map for extraction of a fixed size feature vector using rotated RoIAlign~\cite{Oriented-RCNN2021}. In the second stage of SDR, each feature vector is fed into a series of fully-connected layers with two branches: one for classifying each proposal, and another one for regressing the offsets to the oriented rectangle corresponding to each proposal.
In addition, unlike the regression method used in Arrow R-CNN, we integrate the keypoint detection method based on image segmentation used in Mask R-CNN~\cite{Mask-RCNN2017} to detect the start and end points of the connecting lines with arrowheads, which facilitates the aggregation of the lines in post-processing.
%
%

To better handle the common bus structures in structure diagrams and the corresponding multi-segment polylines between the nodes connected through buses, as shown in Figure~\ref{figure-ownership-organization-examples} (b) and (d), SDR has defined the buses as a special type of detection object in addition to the regular connecting lines as a type of detection object. 
In the post-processing, the regular connecting lines and buses can be aggregated into the multi-segment polylines, which establish the corresponding connection relationship between the nodes.
Figure~\ref{figure-cmp-bus} (b) showed that the SDR can accurately detect all the bus (colored in red) and the attached regular connecting lines (colored in blue). They can be easily aggregated into corresponding multi-segment polylines in post-processing, and the problem of Arrow R-CNN shown in Figure~\ref{figure-cmp-bus} (a) can be avoided.
Compared with the regression method of FR-DETR, our SDR is also very reliable at detecting line segments overlapped by text, as shown in Figure~\ref{figure-cmp-overlap-text} (b), and all the line segments that were not detected in Figure~\ref{figure-cmp-overlap-text} (a) were correctly detected by SDR. 
The shapes of the nodes are relatively limited compared to the connecting lines, so the accuracy of node detection is high for all the above methods, and the bounding boxes of the nodes are shown in green in all the figures.



\subsection{Post-processing} 

The text blocks within a structural diagram are initially detected by an OCR system that utilizes DBNet~\cite{DBnet-2022} and CRNN~\cite{CRNN-2017} and is implemented in PaddleOCR\footnote{https://github.com/PaddlePaddle/PaddleOCR/}. This OCR system was trained on a dataset of Chinese financial announcements. After detection, the text blocks are merged into previously identified nodes or connected lines based on their respective coordinates. For example, if the bounding box of the current text block is located inside the bounding box of a node, the text recognized by OCR is used as the corresponding entity name of that node. Another example is that in an ownership structure diagram if the nearest object to the current text block is a line and the text is a number, that text is extracted as the ownership percentage corresponding to the line. For organization diagrams, text blocks are located only inside nodes and are recognized as names of the departments or occupations.

For each line other than the bus type, its start and end points must be connected to nodes or other lines, so that each line can always find the two bounding boxes closest to its start and end points, respectively. 
For the two bounding boxes that are closest to the start and end of the current line, if both are the node type, the ownership relationship between these two nodes is created based on the ownership percentage corresponding to the current line if they are in an ownership structure diagram, or the subordination relationship between these two nodes is determined based on the current line if they are in an organization structure diagram.
If one of the two bounding boxes closest to the start and end of the current line corresponds to a bus or a line, it is necessary to extend the current line by merging it with the current line until both bounding boxes connected to the current line are of node type.
Not all lines have arrows. For lines with no clear direction, the default direction is top to bottom or left to right.

\subsection{Structured Data Extraction from Diagrams} \label{data_extraction_diagram}

Although the ultimate goal of conducting diagram recognition is to construct financial knowledge graphs, this paper focuses mainly extracting structured data from diagrams, leaving aside tasks such as entity alignment and disambiguation for the time being. 
Each node pair detected and extracted from ownership structure diagrams, along with the corresponding ownership relationship, is output as a relation tuple of \textbf{(Owner, Percentage, Owned)}. Similarly, each node pair detected and extracted from organization structure diagrams, and the corresponding hierarchical relationship, is output as a relation tuple of \textbf{(Supervisor, Subordinate)}.

\section{Semi-automated Two-stage Method for Structure Diagram Annotation} \label{sec_annotate}



Some structure diagrams often have a large number of nodes and dense connecting lines, as shown in Figure~\ref{figure-annotation-examples-1} (a), and annotating all these relatively small object areas also requires more delicate operations, which can be time-consuming and labor-intensive if we rely solely on manual work. To address this problem, we developed a semi-automated, two-stage method to generate high-quality annotations for real-world structure diagrams.

In the first stage, we created an automated tool exploiting the structural properties of structure diagrams to generate structure diagrams for different scenarios and corresponding annotations for training structure diagram recognition models.
In our previous work on extracting information from the textual content of financial announcements, we have built and accumulated a knowledge base of relevant entities of individuals, departments and institutions.
A random integer $n$ can be generated as the number of nodes from a given range, and then $n$ entities can be randomly selected as nodes from the knowledge base according to the type of diagram being synthesized. The number of levels $m$ is randomly chosen from a certain range according to the constraint on the number of nodes $n$, and then the nodes between two adjacent levels are connected in different patterns of one-to-one, one-to-many and many-to-one with certain probabilities, and a small number of shortcut connections between nodes that are not located in adjacent levels can be further randomly generated. Each of these connections corresponds to a specific relationship in a specific type of structure diagram, reflecting the topology in the structure diagram to be synthesized.
A topology generated above is then imported into the visualization tool Graphviz~\cite {Graphviz2003} to automatically draw a corresponding structure diagram according to the settings. In the settings, we can choose the shape and color of the nodes and lines, as well as the orientation and angle of the lines, the font and the position of the text attached to the lines, and whether to use a bus structure to show one-to-many or many-to-one connections. Flexible settings allow the automated tool to synthesize a wide variety of structure diagrams covering a wide range of real-world scenarios (See Appendix~\ref{appendix_syn_examples} for more examples of synthesized structure diagrams).
Graphviz is able to export the synthesized structure diagrams into SVG format, so we can easily obtain the coordinates of each object in the diagrams for corresponding annotations, and finally convert them into DOTA format~\cite{DOTA2021} as training data. Then a preliminary model can be trained based on the automatically synthesized training data.

In the second stage, the preliminary model can be used to automatically annotate the real-world structure diagrams extracted from financial announcements. Our experiments in Section~\ref{sec_datasets} show that the preliminary model usually has a reasonably good performance. A typical example in Figure~\ref{figure-annotation-examples-1} (b) shows the results of the automatic annotation of a diagram using the preliminary model, where only a very small number of short vertical line segments are not correctly detected, and generally, we only need to make a few corrections to the automatic annotation results (See Appendix~\ref{appendix_auto_examples} for more examples of structure diagrams automatically annotated by the preliminary model). The manual correction tool first converts the auto-annotated DOTA data into the COCO format, then imports it into the COCO Annotator for correction, and finally converts it back to the DOTA format.

\section{Experiments}\label{sec_experiments}


\subsection{Evaluation Metrics} 


Average Precision (AP) and mean Average Precision (mAP) are the most common metrics used to evaluate object detection~\cite{Faster-R-CNN2015}. 
Arrow key points do not participate in the mAP calculation because they are predicted by a separate head that is different from nodes, buses, and lines.
We also use $Precision/Recall/F1$ as metrics for object detection, which are commonly used in flowchart recognition~\cite{Arrow-RCNN2021,FR-DETR-2022}. The IoU threshold for the above metrics is 50$\%$.

However, in the task of structure diagram recognition, object detection is only one part of the process, and the ultimate goal is to obtain the topology of the structure diagram and extract the tuples of structured data for import into the knowledge base.
Therefore, a method cannot only be measured in terms of object detection, but must also be evaluated on the structured data extracted through aggregation and post-processing.
For an ownership relationship tuple \textbf{(Owner, Percentage, Owned)} or a subordinate relationship tuple \textbf{(Supervisor, Subordinate)}, a tuple is correct only if all elements contained in that tuple are correctly extracted. By checking the extracted tuples in each structure diagram, the $Precision/Recall/F1$ of the extracted tuples can be counted as metrics for evaluating structured data extraction.

\subsection{Datasets} \label{sec_datasets}

\subsubsection{Synthesized Dataset}

Based on the observation of the real data distribution, we automatically synthesized and annotated 8050 ownership structure diagrams and 4450 organization structure diagrams using the automated tool introduced in section~\ref{sec_annotate}. Based on this synthesized dataset, we can train a preliminary SDR model for structure diagram recognition.

\subsubsection{Real-world Benchmark Dataset}

Following the method presented in section~\ref{diagram_detection}, we first used layout analysis to automatically detect and extract some structure diagrams from publicly disclosed Chinese financial announcements such as prospectuses, and then invited several financial professionals to manually review and finally select 2216 ownership structure diagrams and 1750 organization diagrams. 
The principle is to cover as wide a range of different layout structures and styles as possible, including nodes of different shapes, colors, and styles, as well as connecting lines of different patterns, directions, and angles, in an attempt to maintain the diversity and complexity of the structure diagrams in a real-world scenario.
Then, using the two-stage method presented in section \ref{sec_annotate}, we applied the preliminary SDR model trained on the synthesized dataset to automatically annotate all diagrams in the above real-world dataset, and then manually corrected the automatic annotations to create the industry's first structure diagram benchmark.

Comparing the results of automatic annotation with the results after manual corrections, the preliminary SDR model showed a fairly good performance, as shown in Table~\ref{preliminary_model}, and only a small number of corrections are needed. So actual experiments verify that the two-stage method can significantly improve efficiency and reduce costs.

As introduced in Section~\ref{sec_detection_line_node}, Arrow RCNN, FR-DETR, and our SDR have different definitions of connecting lines, so the corresponding annotations are different as well. However, we can easily convert the annotation data used in our SDR to the annotation data used in Arrow R-CNN and FR-DETR automatically.
A preliminary Arrow R-CNN model and a preliminary FR-DETR model were also trained on the synthesized dataset, respectively, and used to automatically annotate all the above real dataset.
Table~\ref{preliminary_model} shows that the preliminary Arrow R-CNN model is poor. However, the preliminary FR-DETR model is also reasonably good.

\begin{table}
  \centering
  \caption{Evaluation of the different preliminary models on the entire benchmark dataset}\label{preliminary_model}
  \tiny
  \setlength{\tabcolsep}{0.5mm}
  \begin{tabular}{c|c|c|c|c|c|c|c|c|c}
    \Xhline{1pt}
    \multicolumn{2}{c}{Datasets} & \multicolumn{4}{|c}{Ownership} & \multicolumn{4}{|c}{Organization} \\
    \hline
    {Category}& {Metrics(\%)} & {\makecell[c]{Arrow \\ R-CNN}} & {FR-DETR} & {\makecell[c]{Our SDR \\ (R50)}} & {\makecell[c]{Our SDR \\ (Swin-S)}} & {\makecell[c]{Arrow \\ R-CNN}} & {FR-DETR} & {\makecell[c]{Our SDR \\ (R50)}} & {\makecell[c]{Our SDR \\ (Swin-S)}} \\
    \hline
    \multirow{4}*{Node}
    & Precision & \textbf{97.5} & 96.1 & 92.7 & 92.5 & 97.2 & 95.6 & 98.9 & \textbf{99.4} \\
    & Recall & 98.6 & 98.9 & \textbf{99.3} & 99.2 & 98.1 & 99.7 & \textbf{99.9} & 99.8 \\
    & F1 & \textbf{98.0} & 97.5 & 95.9 & 95.7 & 97.6 & 97.6 & 99.4 & \textbf{99.6} \\
    \cline{2-10}
    & AP & 96.8 & 95.3 & \textbf{98.8} & 98.7 & 96.2 & 98.7 & 98.9 & \textbf{99.9} \\
    \hline
    \multirow{4}*{Line}
    & Precision & 49.0 & 71.9 & 79.0 & \textbf{83.8} & 52.7 & 82.0 & 85.7 & \textbf{88.6} \\
    & Recall & 53.8 & 76.5 & 83.0 & \textbf{88.8} & 54.8 & \textbf{86.7} & 81.8 & 82.2 \\
    & F1 & 51.3 & 74.1 & 81.0 & \textbf{86.2} & 53.7 & 84.3 & 83.7 & \textbf{85.3} \\
    \cline{2-10}
    & AP & 47.4 & 67.0 & 77.9 & \textbf{85.3} & 48.3 & 81.3 & 79.3 & \textbf{81.0} \\
    \hline
    \multirow{4}*{Bus}
    & Precision & N/A & N/A & 66.9 & \textbf{75.5} & N/A & N/A & 54.6 & \textbf{70.2} \\
    & Recall & N/A & N/A & 80.6 & \textbf{89.4} & N/A & N/A & 94.4 & \textbf{98.2} \\
    & F1 & N/A & N/A & 73.1 & \textbf{81.9} & N/A & N/A & 69.2 & \textbf{81.9} \\
    \cline{2-10}
    & AP & N/A & N/A & 72.9 & \textbf{84.5} & N/A & N/A & 91.4 & \textbf{97.0} \\
    \hline
    \multicolumn{2}{c|}{mAP} & 72.1 & 81.2 & 83.2 & \textbf{89.5} & 72.3 & 90.0 & 89.9 & \textbf{92.3} \\
    \hline
    \multirowcell{4}{Arrow keypoints}
    & Precision & 49.5 & N/A & 86.8 & \textbf{91.6} & 41.3 & N/A & 74.9 & \textbf{75.3} \\
    & Recall & 52.4 & N/A & 95.5 & \textbf{96.8} & 44.8 & N/A & \textbf{96.1} & 95.6 \\
    & F1 & 50.9 & N/A & 90.9 & \textbf{94.1} & 43.0 & N/A & \textbf{84.2} & \textbf{84.2} \\
    \cline{2-10}
    & AP & 46.9 & N/A & 83.0 & \textbf{88.7} & 39.6 & N/A & 71.6 & \textbf{72.2} \\
    
    \Xhline{1pt}
  \end{tabular}
\end{table}

\subsection{Implementation of Baselines and our SDR}


As discussed in the previous sections, Arrow R-CNN and FR-DETR are the methods closest to our work, so they are selected as the baselines.
%
Arrow R-CNN mainly extended Faster R-CNN for better handwritten flowchart recognition. 
It is not yet open source, so we modify the Faster R-CNN model obtained from Detectron2\footnote{https://github.com/facebookresearch/detectron2} to reproduce Arrow R-CNN according to the description in the original paper~\cite{Arrow-RCNN2021}. 
FR-DETR simply merged DETR and LETR into a multi-task model, and the experiments~\cite{FR-DETR-2022} showed that the multi-task FR-DETR performed slightly worse than the single-task of DETR in symbol detection and slightly worse than the single-task LETR in line segment detection, respectively. So even though FR-DETR is not open source, we can use separate DETR to get FR-DETR's upper bounds on node detection performance, and separate LETR to get FR-DETR's upper bounds on line segment detection performance. Our DETR codes are from the official implementation in Detectron2, and our LETR codes are from the the official implementation of the paper~\cite{LETR-2021}.
Our SDR model mainly extended Oriented R-CNN implementation in MMRotate~\cite{mmrotate-zhou2022}. We tried two different backbones based on ResNet50 and Swin-Transformer-Small, referred to in all the tables as \textbf{R50} and \textbf{Swin-S}, respectively.
See Appendix~\ref{model_setting} for more implementation details.


\subsection{Evaluation on Real-world Benchmark}

\begin{table}
  \tiny
  \centering
  \caption{Evaluation on the recognition of ownership structure diagrams (Pre: Preliminary models, FT: Fine-tuned models)}\label{ownership_evaluation}
  \setlength{\tabcolsep}{0.8mm}
  \begin{tabular}{c|c|c c|c c|c c|c c}
    \Xhline{1pt}
     \multirow{2}*{Category}& \multirow{2}*{Metrics(\%)} & \multicolumn{2}{c}{Arrow R-CNN} & \multicolumn{2}{|c}{FR-DETR} & \multicolumn{2}{|c}{Our SDR(R50)} & \multicolumn{2}{|c}{Our SDR(Swin-S)}\\
    \cline{3-10}
    & & Pre & FT & Pre & FT & Pre & FT & Pre & FT\\
    \hline
    \multirow{4}*{Node}
    & Precision & 97.6 & 98.7 & 96.9 & \textbf{99.5} & 93.3 & 99.0 & 92.9 & 99.1\\
    & Recall & 98.8 & 99.0 & 98.8 & \textbf{99.8} & 99.7 & \textbf{99.8} & 99.5 & \textbf{99.8}\\
    & F1 & 98.2 & 98.8 & 97.8 & \textbf{99.6} & 96.4 & 99.4 & 96.1 & 99.4 \\
    \cline{2-10}
    & AP & 96.7 & 98.9 & 96.3 & 98.8 & 98.8 & \textbf{98.9} & 98.7 & \textbf{98.9} \\
    \hline
    \multirow{4}*{Line}
    & Precision & 50.9 & 67.6 & 73.2 & 88.5 & 79.4 & 92.8 & 84.5 & \textbf{93.7} \\
    & Recall & 53.6 & 70.7 & 76.7 & 90.3 & 83.8 & 98.3 & 90.5 & \textbf{98.8} \\
    & F1 & 52.2 & 69.1 & 74.9 & 89.4 & 81.5 & 95.5 & 87.4 & \textbf{96.2} \\
    \cline{2-10}
    & AP & 48.1 & 65.5 & 70.9 & 87.8 & 78.3 & 97.3 & 86.5 & \textbf{98.2} \\
    \hline
    \multirow{4}*{Bus}
    & Precision & N/A & N/A & N/A & N/A & 66.7 & 79.8 & 74.3 & \textbf{85.3} \\
    & Recall & N/A & N/A & N/A & N/A & 81.6 & 97.7 & 90.1 & \textbf{98.4} \\
    & F1 & N/A & N/A & N/A & N/A & 73.4 & 87.8 & 81.4 & \textbf{91.4} \\
    \cline{2-10}
    & AP & N/A & N/A & N/A & N/A & 73.2 & 96.5 & 85.0 & \textbf{97.6} \\
    \hline
    \multicolumn{2}{c|}{mAP} 
    & 72.4 & 82.2 & 83.6 & 93.3 & 83.4 & 97.6 & 90.1 & \textbf{98.2} \\
    \hline
    \multirowcell{4}{Arrow keypoints}
    & Precision & 50.4 & 54.8 & N/A & N/A & 86.7 & 96.9 & 91.3 & \textbf{97.0} \\
    & Recall & 49.1 & 57.6 & N/A & N/A & 96.0 & \textbf{98.7} & 97.0 & 98.5 \\
    & F1 & 47.2 & 56.2 & N/A & N/A & 91.1 & \textbf{97.8} & 94.1 & 97.7 \\
    \cline{2-10}
    & AP & 44.0 & 52.7 & N/A & N/A & 83.7 & \textbf{95.1} & 88.9 & 95.0 \\
    
    \Xhline{1pt}
  \end{tabular}
\end{table}

\begin{table}
  \tiny
  \centering
  \caption{Evaluation on the recognition of organization structure diagrams (Pre: Preliminary models, FT: Fine-tuned models)}\label{organization_evaluation}
  \setlength{\tabcolsep}{0.8mm}
  \begin{tabular}{c|c|c c|c c|c c|c c}
    \Xhline{1pt}
     \multirow{2}*{Category}& \multirow{2}*{Metrics(\%)} & \multicolumn{2}{c}{Arrow R-CNN} & \multicolumn{2}{|c}{FR-DETR} & \multicolumn{2}{|c}{Our SDR(R50)} & \multicolumn{2}{|c}{Our SDR(Swin-S)}\\
    \cline{3-10}
    & & Pre & FT & Pre & FT & Pre & FT & Pre & FT\\
    \hline
    \hline
    \multirow{4}*{Node}
    & Precision & 96.7 & 98.9 & 98.0 & 99.6 & 99.4 & \textbf{99.8} & 99.6 & 99.7\\
    & Recall & 98.9 & 99.3 & 98.8 & \textbf{99.9} & \textbf{99.9} & \textbf{99.9} & 99.8 & \textbf{99.9}\\
    & F1 & 97.8 & 99.7 & 98.4 & 99.7 & 99.6 & \textbf{99.8} & 99.7 & \textbf{99.8} \\
    \cline{2-10}
    & AP & 96.5 & 98.7 & 97.9 & 99.0 & 98.8 & \textbf{99.0} & 98.9 & \textbf{99.0} \\
    \hline
    \multirow{4}*{Line}
    & Precision & 51.5 & 77.4 & 89.0 & 95.5 & 87.3 & 95.9 & 88.0 & \textbf{98.4} \\
    & Recall & 53.2 & 73.5 & 86.4 & \textbf{96.8} & 79.2 & 95.4 & 79.9 & 95.3 \\
    & F1 & 52.3 & 75.4 & 87.7 & 96.1 & 83.1 & 95.6 & 83.8 & \textbf{96.8} \\
    \cline{2-10}
    & AP & 49.0 & 68.3 & 81.6 & 94.0 & 77.4 & \textbf{95.0} & 78.0 & \textbf{95.0} \\
    \hline
    \multirow{4}*{Bus}
    & Precision & N/A & N/A & N/A & N/A & 53.8 & 90.0 & 69.9 & \textbf{95.3} \\
    & Recall & N/A & N/A & N/A & N/A & 93.9 & 98.3 & 97.9 & \textbf{98.5} \\
    & F1 & N/A & N/A & N/A & N/A & 68.4 & 94.0 & 81.6 & \textbf{96.9} \\
    \cline{2-10}
    & AP & N/A & N/A & N/A & N/A & 91.1 & \textbf{98.0} & 96.4 & 97.9 \\
    \hline
    \multicolumn{2}{c|}{mAP} 
    & 72.8 & 83.5 & 89.8 & 96.5 & 89.1 & \textbf{97.3} & 91.1 & \textbf{97.3} \\
    \hline
    \multirowcell{4}{Arrow keypoints}
    & Precision & 42.0 & 51.0 & N/A & N/A & 70.3 & \textbf{97.7} & 72.5 & 96.3 \\
    & Recall & 44.3 & 54.2 & N/A & N/A & 95.9 & \textbf{99.6} & 94.3 & \textbf{99.6} \\
    & F1 & 43.1 & 52.6 & N/A & N/A & 81.1 & \textbf{98.6} & 82.0 & 97.9 \\
    \cline{2-10}
    & AP & 40.2 & 48.1 & N/A & N/A & 67.7 & \textbf{96.7} & 68.6 & 95.7 \\
    
    \Xhline{1pt}
  \end{tabular}
\end{table}

\begin{table}
  \tiny
  \centering
  \caption{Evaluation of extracting structured data from the ownership structure diagrams (Pre: Preliminary models, FT: Fine-tuned models)}\label{ownership_structured_data}
  \setlength{\tabcolsep}{0.8mm}
  \begin{tabular}{c|c c|c c|c c|c c}
    \Xhline{1pt}
     \multirow{2}*{Metrics(\%)} & \multicolumn{2}{c}{Arrow R-CNN} & \multicolumn{2}{|c}{FR-DETR} & \multicolumn{2}{|c}{Our SDR(R50)} & \multicolumn{2}{|c}{Our SDR(Swin-S)}\\
    \cline{2-9}
    & Pre & FT & Pre & FT & Pre & FT & Pre & FT\\
    \hline
    Precision & 45.6 & 61.2 & 78.7 & 85.6 & 80.2 & \textbf{91.5} & 85.2 & 91.4\\
    Recall & 47.9 & 64.8 & 72.2 & 82.8 & 75.6 & 87.0 & 81.3 & \textbf{87.6}\\
    F1 & 46.7 & 62.9 & 75.3 & 84.2 & 77.8 & 89.2 & 83.2 & \textbf{89.5} \\
    \Xhline{1pt}
  \end{tabular}
\end{table}

\begin{table}
  \tiny
  \centering
  \caption{Evaluation of extracting structured data from the organization structure diagrams (Pre: Preliminary models, FT: Fine-tuned models)}\label{organization_structured_data}
  \setlength{\tabcolsep}{0.8mm}
  \begin{tabular}{c|c c|c c|c c|c c}
    \Xhline{1pt}
     \multirow{2}*{Metrics(\%)} & \multicolumn{2}{c}{Arrow R-CNN} & \multicolumn{2}{|c}{FR-DETR} & \multicolumn{2}{|c}{Our SDR(R50)} & \multicolumn{2}{|c}{Our SDR(Swin-S)}\\
    \cline{2-9}
    & Pre & FT & Pre & FT & Pre & FT & Pre & FT\\
    \hline
    Precision & 46.5 & 66.1 & 81.4 & 90.9 & 81.8 & \textbf{91.9} & 82.3 & 91.8 \\
    Recall & 50.6 & 69.0 & 78.5 & 86.6 & 76.4 & 87.4 & 77.9 & \textbf{87.5} \\
    F1 & 48.5 & 67.5 & 79.9 & 88.7 & 79.0 & \textbf{89.6} & 80.0 & \textbf{89.6} \\
    \Xhline{1pt}
  \end{tabular}
\end{table}

\begin{table}
  \centering
  \caption{Parameter number and inference time of different models.}\label{params_inference}
  \tiny
  \setlength{\tabcolsep}{1mm}
  \begin{tabular}{c|c|c}
    \Xhline{1pt}
    {Network} & {params} & {seconds per image} \\
    \hline
    Arrow R-CNN & 59.0M & 0.18 \\
    FR-DETR & 59.5M & 1.22 \\
    Our SDR(R50) & 59.1M & \textbf{0.16} \\
    Our SDR(Swin-S) & 83.8M & 0.21 \\
    \Xhline{1pt}
  \end{tabular}
\end{table}


After shuffling the real-world benchmark, we selected 1772 diagrams as the training set and 444 diagrams as the test set for the ownership structure diagrams, and 1400 diagrams as the training set and 350 diagrams as the test set for the organization structure diagrams.

For the input size of diagram images, the longer side of the image is scaled to 1024 and the shorter side is subsequently scaled according to the aspect ratio of the original image before being input to the model. For all models in the experiments, we followed the above configuration.

Table~\ref{ownership_evaluation} and Table~\ref{organization_evaluation} showed the evaluations performed on the test set for the ownership structure diagrams and the test set for the organization structure diagrams, respectively. As discussed in Section~\ref{sec_detection_line_node}, the shapes of the nodes are relatively limited compared to the connecting lines, so the performance of node detection is good for all the above methods, and we focus on the detection of connecting lines in this section. 
Although the synthesized data tried to capture the real data distribution as much as possible, there are always some complex and special cases in the real data that differ from the regular scenarios, which inevitably lead to some differences between the two. Therefore, fine-tuning on real training set can allow the model to better adapt to the real data distribution and thus further improve the performance of the model, and the experimental results in Table~\ref{ownership_evaluation} and Table~\ref{organization_evaluation} showed the effectiveness of fine-tuning.

Sch{\"a}fer et al.~\cite{Arrow-RCNN2021} also observed that when some connecting lines are very close together in handwritten flowcharts, their horizontal bounding boxes overlap each other to a large extent, so they tried to increase the IoU threshold of NMS from 0.5 used in Faster R-CNN to 0.8 to improve the detection of lines with overlapping bounding boxes in Arrow R-CNN.
After increasing the threshold, the detection performance of the connection lines did improve significantly, with F1 increasing from $51.8\%$ to $69.1\%$ on the ownership test set and from $57.7\%$ to $73.5\%$ on the organization test set, but Arrow R-CNN is still difficult to correctly detect many common connecting lines whose horizontal rectangular bounding boxes completely overlap each other, as shown in Figure~\ref{figure-cmp-inclined} (a) and Figure~\ref{figure-cmp-bus} (a).
Therefore, the performance of the Arrow R-CNN in the detection of connecting lines is still not so good.
%

FR-DETR does not do a good job of detecting curves, but curves occur in a relatively small percentage of the current dataset of structure diagrams, so this has less impact on its overall performance.
FR-DETR only detects each individual straight line segment as an object, instead of detecting each complex ployline containing multiple line segments as an object like Arrow R-CNN. Therefore, its task is simpler than Arrow R-CNN, and the corresponding performance is much better.
As shown in Figure~\ref{figure-cmp-overlap-text} (a), there are often text overlays on connecting lines in ownership structure diagrams, and LETR in FR-DETR is not very robust and can cause many connecting lines to be missed. Therefore, as shown in Table~\ref{ownership_evaluation}, FR-DETR's performance is not as good as SDRs. However, for organization structure diagrams, the connecting lines are generally not covered by text, so FR-DETR's detection of connecting lines by FR-DETR is not affected, and its performance is very close to that of SDRs, as shown in Table~\ref{organization_evaluation}.

Consistent with Table~\ref{preliminary_model}, Table~\ref{ownership_evaluation} and Table~\ref{organization_evaluation} show that the SDR (Swin-S) has a significant advantage over the SDR (R50) on the preliminary models before fine tuning on real data. This may be due to the fact that Swin-Transformer has better feature extraction and representation capabilities, and thus better generalization capabilities. The advantage of the SDR (Swin-S) over the SDR (R50) is relatively small after fine-tuning on real data. 

We also examined the SDR failure cases, which are typically due to very short lines connected to the bus, complex structures containing intersecting lines, and interference from dashed line boxes.
See Appendix~\ref{failure_cases} for some examples.
%

%
%
To facilitate the extraction of topology and structured data, our SDR defines buses as a special type of object that allows for easier and more reliable aggregation in post-processing.
The relationship tuples obtained from structured data extraction were evaluated on the test sets of the ownership structure diagrams and the organization structure diagrams, respectively, according to the method introduced in Section~\ref{data_extraction_diagram}. Table~\ref{ownership_structured_data} and Table~\ref{organization_structured_data} show that SDR is also better than FR-DETR, and Arrow R-CNN is still the worst, and the the fine-tuning also has significant effects.

Table~\ref{params_inference} compares the number of parameters and inference speed of the above models.
SDR (R50) has the fastest speed and its number of parameters is comparable to Arrow R-CNN, but its recognition performance is much better than Arrow R-CNN.
The number of parameters of FR-DETR is comparable to that of SDR (R50), and FR-DETR's overall recognition performance is also not far from that of SDR (R50), but SDR (R50)'s inference speed is more than 7 times FR-DETR's. FR-DETR's speed bottleneck is mainly the LETR part.
SDR (Swin-S) has the largest number of parameters and the best recognition performance, and its inference speed is $30\%$ slower than SDR(R50), and but still more than $4$ times faster than FR-DETR.

All the above experimental results show that the SDR method proposed in this paper has significantly improved compared to the previous methods, especially the detection of various connecting lines, which in turn leads to an improvement in the structured data extraction corresponding to the connection relationships.

\section{Conclusion}\label{sec_conclusion}

In this paper, we proposed a new method for structure diagram recognition that can better detect various complex connecting lines. We also developed a two-stage method to efficiently generate high-quality annotations for real-world diagrams, and constructed the industry's first structure diagram benchmark from real financial announcements. Empirical experiments validated the significant performance advantage of our proposed methods over existing methods. In the future, we plan to extend the methods in this paper and apply it to the recognition of more types of diagrams, such as process diagrams.

\bibliographystyle{splncs04}
\bibliography{ref}






\section{Appendix}

\subsection{Examples of Synthesized Structure Diagrams}\label{appendix_syn_examples}

\begin{figure}
\centering
\includegraphics[width=\textwidth]{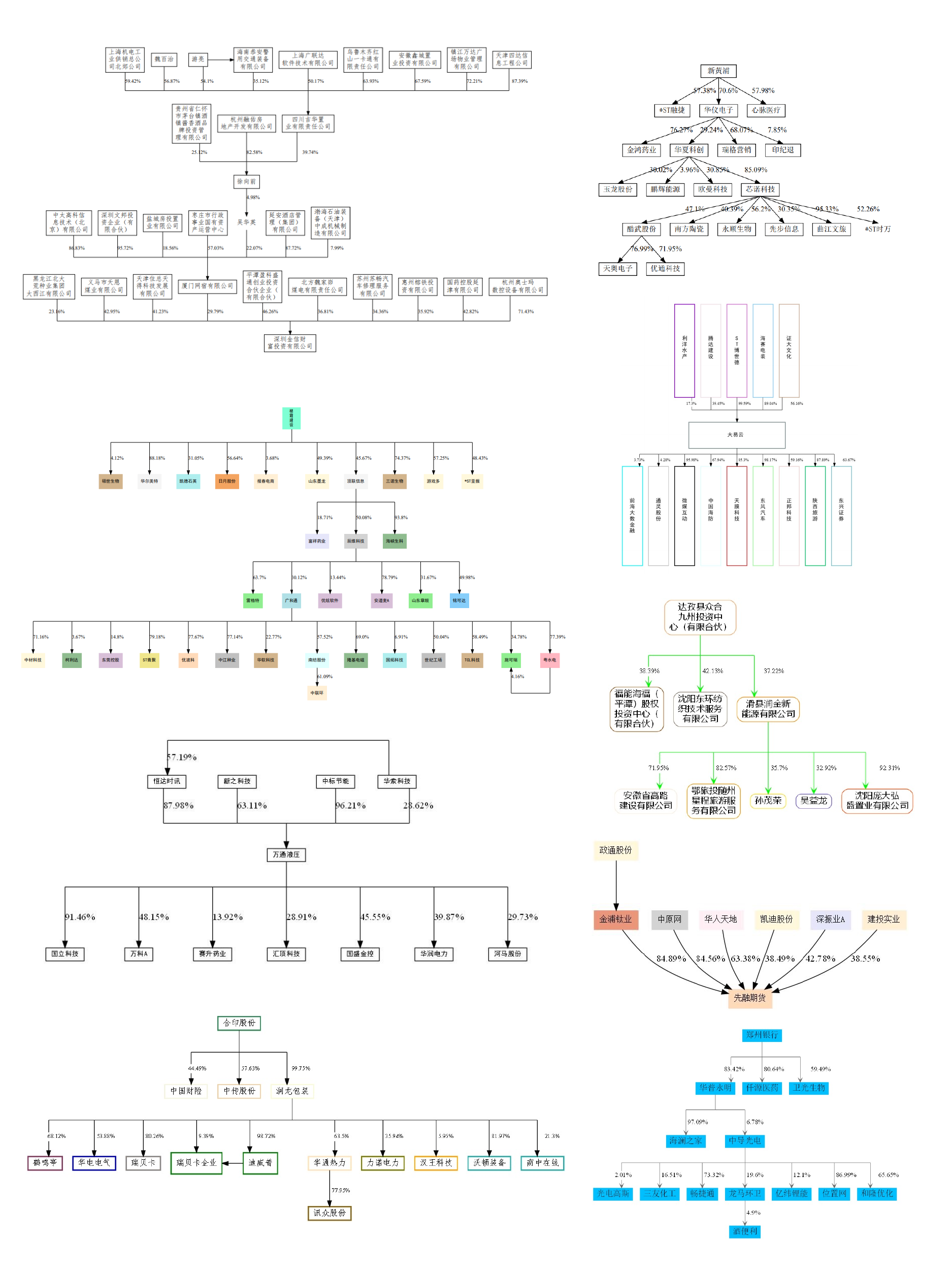}
\caption{Some examples of Synthesized Ownership Structure Diagrams} \label{figure_syn_ownership_examples}
\end{figure}

\begin{figure}
\centering
\includegraphics[width=\textwidth]{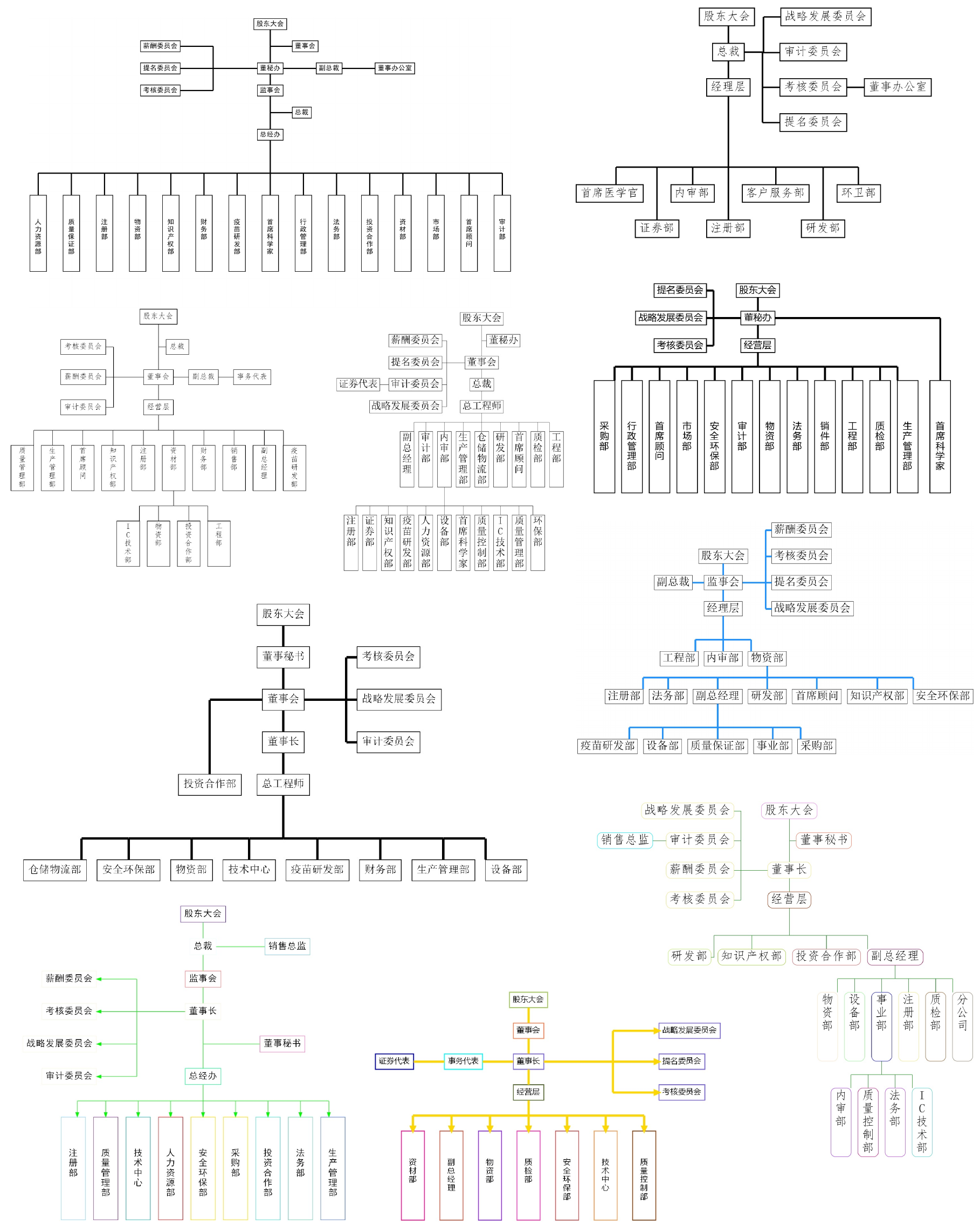}
\caption{Some examples of Synthesized Organization Structure Diagrams} \label{figure_syn_organization_examples}
\end{figure}

Figure~\ref{figure_syn_ownership_examples} and Figure~\ref{figure_syn_organization_examples} show some examples of synthesized ownership structure diagrams and organization structure diagrams, respectively.

\subsection{Examples of Structure Diagrams Automatically Annotated by the Preliminary SDR Models}\label{appendix_auto_examples}

\begin{figure}
\centering
\includegraphics[width=\textwidth]{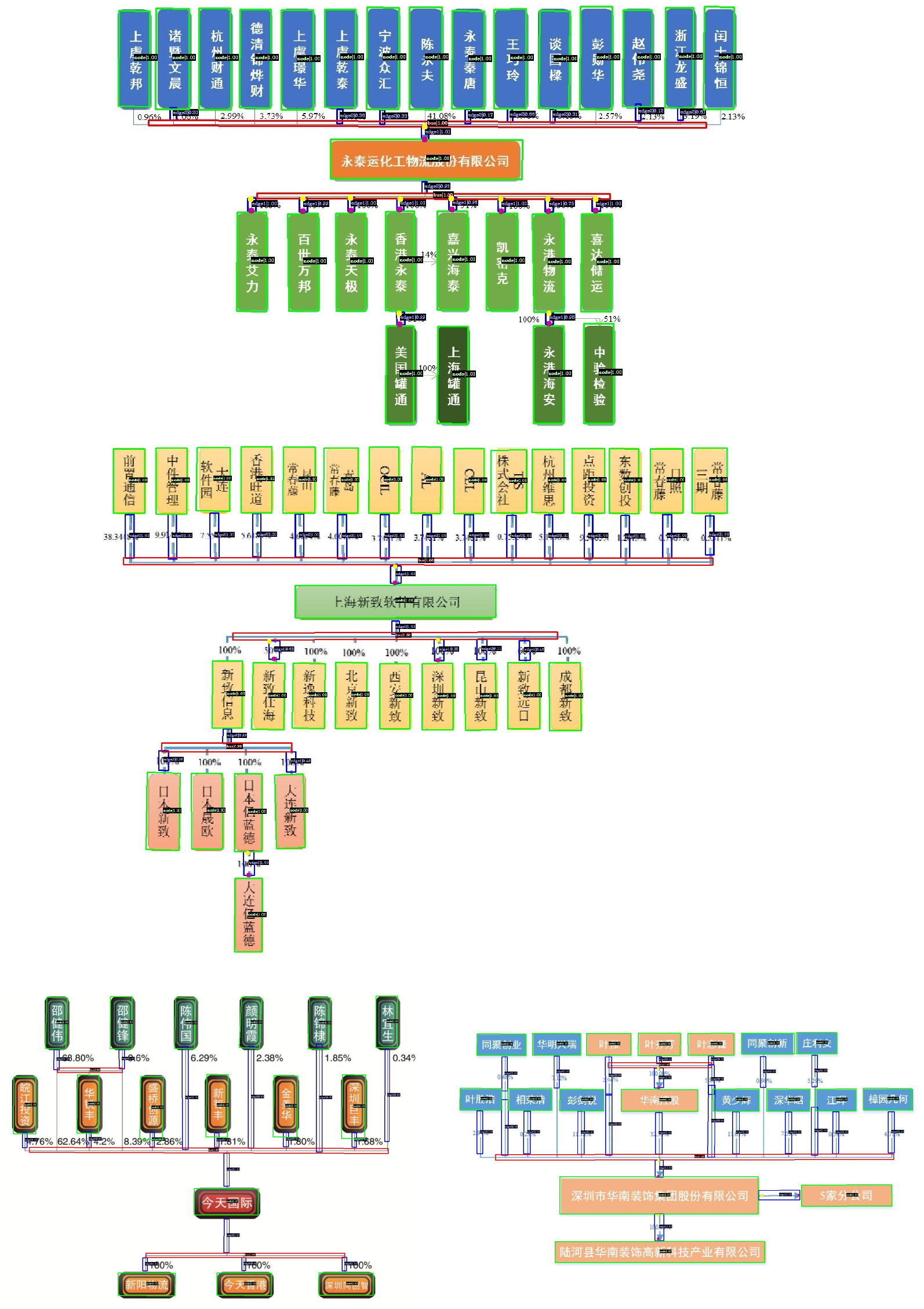}
\caption{Some examples of Ownership Structure Diagrams Automatically Annotated by the Preliminary SDR Model} \label{figure_auto_ownership_examples}
\end{figure}

\begin{figure}
\centering
\includegraphics[width=\textwidth]{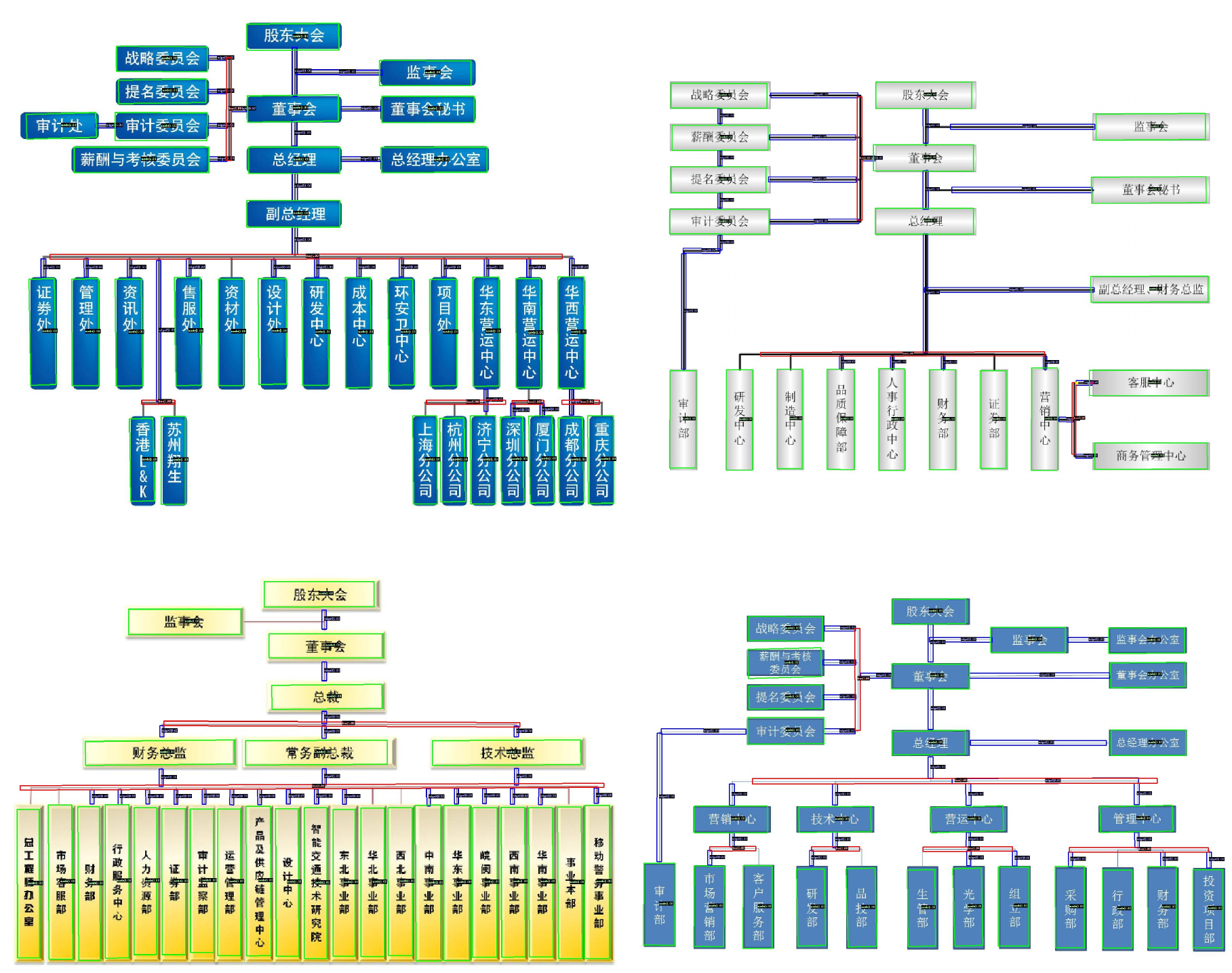}
\caption{Some examples of Organization Structure Diagrams Automatically Annotated by the Preliminary SDR Model} \label{figure_auto_organization_examples}
\end{figure}

Figure~\ref{figure_auto_ownership_examples} and Figure~\ref{figure_auto_organization_examples} show some examples of ownership structure diagrams and organization structure diagrams, respectively, automatically annotated by the preliminary SDR models.

\subsection{Implementation details for Arrow R-CNN, FR-DETR and SDR}\label{model_setting}

\subsubsection{Arrow R-CNN}
The backbone is ResNet50 with FPN, and is initialized with the ImageNet-based pre-training model~\footnote{https://github.com/facebookresearch/detectron2}.
The model was trained using Momentum SGD as the optimizer, with a batch size of 2, a maximum number of iterations of 38000, an initial learning rate of 0.0025, and dividing by 10 at the 32000th and 36000th iteration. Horizontal and vertical flips and rotations were used for data augmentation during training.

\subsubsection{FR-DETR}
For DETR, the backbone is ResNet50 with pretrained weights~\footnote{https://github.com/facebookresearch/detr}. The model was trained with a batch size of 2 and a total of 300 epochs, and the learning rate was divided by 10 at the 200th epochs. The model was optimized using AdamW with the learning rate of 1e-04, and the weight decay of 1e-04. Only horizontal flips were used for data augmentation during training.

For LETR, The backbone is ResNet50~\footnote{https://github.com/mlpc-ucsd/LETR}. The model was trained with a batch size of 1, and 25 epochs for focal-loss fine-tuning on the pre-training model. The model was optimized using AdamW with the learning rate of 1e-05, and the weight decay of 1e-4.
Horizontal and vertical flips were used for data augmentation during training.

\subsubsection{SDR}

The codes are modified and extended from MMRotate~\footnote{https://github.com/open-mmlab/mmrotate}.
The model was trained with a batch size of 1 and a total of 12 epochs, and the learning rate was divided by 10 at the 8th and 11th epochs. Horizontal and vertical flips and rotations were used for data augmentation.

One option for the backbone is ResNet50, and the network was optimized using the SGD algorithm with a momentum of 0.9, a weight decay of 0.0001, and an initial learning rate set to 1.25e-03. 

Another backbone option is Swin-Transformer-small, and the network was optimized using AdamW with the learning rate of 2.5e-05, and the weight decay of 0.05. 

Our SDR provided seven customized anchor box aspect ratios (0.02, 0.1, 0.5, 1.0, 2.0, 4.0, 10.0) to accommodate different types of objects with different sizes and shapes in structure diagrams.

\subsection{Examples of SDR Failure Cases}\label{failure_cases}

\begin{figure}
\centering
\includegraphics[width=\textwidth]{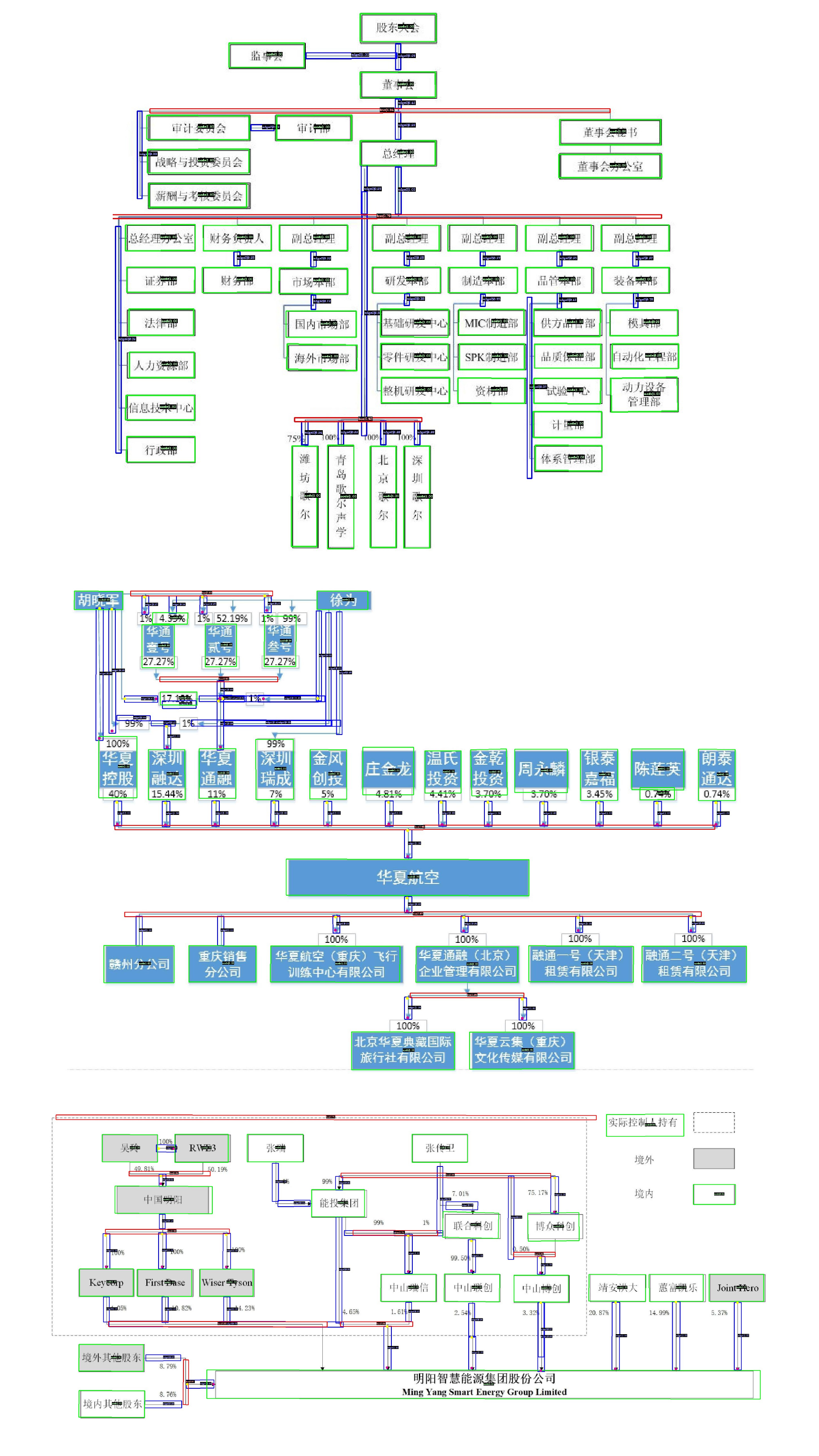}
\caption{Some examples of SDR failure cases of detecting connecting lines} \label{figure_failure_case_examples}
\end{figure}

Figure~\ref{figure_failure_case_examples} shows some examples of SDR failure cases of detecting connecting lines.

\end{document}